# Optimizing Classification of Infrequent Labels by Reducing Variability in Label Distribution

Ashutosh Agarwal

School of Business

Stevens Institute of Technology

Hoboken, USA

ashutoshagarwal198@gmail.com

*Abstract*—This paper presents a novel solution, LEVER, designed to address the challenges posed by underperforming infrequent categories in Extreme Classification (XC) tasks. Infrequent categories, often characterized by sparse samples, suffer from high label inconsistency, which undermines classification performance. LEVER mitigates this problem by adopting a robust Siamese-style architecture, leveraging knowledge transfer to reduce label inconsistency and enhance the performance of One-vs-All classifiers. Comprehensive testing across multiple XC datasets reveals substantial improvements in the handling of infrequent categories, setting a new benchmark for the field. Additionally, the paper introduces two newly created multi-intent datasets, offering essential resources for future XC research.

## I. INTRODUCTION

In contemporary machine learning, the cost and effort required to obtain labeled data remain significant challenges, especially in domains like medical diagnostics, natural language understanding, and computer vision, where expert annotation is resource-intensive. Semi-supervised learning (SSL) emerges as a practical and effective paradigm to address this issue by leveraging the vast quantities of readily available unlabeled data in conjunction with a relatively small set of labeled examples. Unlike traditional supervised methods that rely solely on annotated datasets, SSL techniques aim to enhance learning models by extracting meaningful patterns from both labeled and unlabeled sources. This approach not only reduces dependency on human supervision but also improves the generalization capability of models, especially when deployed in real-world, data-scarce environments.

A key area of advancement within SSL involves graphbased methods, which construct a graph to represent data instances as nodes and encode relationships among them via weighted edges. This graph-centric representation captures the underlying manifold structure of the data, thereby enabling efficient propagation of label information across the graph. Specifically, similar instances are linked together, facilitating the diffusion of known labels to unannotated samples based on their proximity in the graph structure. The general optimization problem in graph-based SSL can be formulated as follows:

$$\min_u J(u) = \sum_{i,j} w_{ij} L(u(x_i), u(x_j)), \qquad (1)$$

where $w_{ij}$ measures the similarity between data points $x_i$ and $x_j$, and L represents a loss function quantifying the dissimilarity between their assigned labels. When L corresponds to the $\ell_p$ norm and the known labels are enforced such that $u(x_i) = y_i$ for the labeled index set $i \in [l]$, the method is referred to as $p$-Laplacian learning. In the special case where $p = 2$, this reduces to the classical Laplace learning model.

Despite their elegance and simplicity, standard graph Laplacian methods suffer from a critical limitation: in scenarios where the graph is extremely large or when labeled data is sparse, the label propagation process can become degenerate. Specifically, the optimization often leads to trivial or overly smooth solutions where the predicted labels across unlabeled nodes converge to nearly constant values. This phenomenon is attributed to the loss of discriminative information during long-range diffusion, effectively causing the model to forget the contextual information embedded in the original labels. To combat this issue, researchers have proposed higher-order Laplacians, anisotropic diffusion models, and more recently, Poisson learning techniques, which introduce source terms to reinvigorate label variation across the graph.

Building on these developments, we propose a novel method called *Variance-Enhanced Poisson Learning (VPL)*, which introduces a simple yet powerful regularization term to the Poisson learning framework. The primary intuition behind VPL is to enforce label diversity by penalizing overly uniform predictions, thereby mitigating the degeneracy observed in conventional diffusion-based models. This is achieved through the incorporation of a variance-amplifying penalty in the objective function, which explicitly encourages the label function to retain informative spread across unlabeled data.

The major contributions of this work are as follows:

- We introduce VPL, a new framework for graph-based SSL that improves performance in settings with limited labeled data by counteracting solution degeneracy.
- We present two specific instantiations of VPL: VLaplace, which extends the traditional Laplacian framework, and V-Poisson, which builds upon the Poisson learning formulation by integrating a variance-enhancing term.
- We incorporate the VPL framework into graph neural networks, resulting in the Variance-Enhanced Graph Poisson Network (V-GPN), thus enabling its application to large-scale and deep graph-based architectures.
- We conduct a rigorous theoretical analysis in both discrete (finite-sample) and continuous (variational) settings,

which leads to the derivation of a novel weighted $p$ Poisson equation that generalizes classical graph SSL formulations.

Through extensive experiments on benchmark datasets, we demonstrate that VPL significantly improves the performance and robustness of semi-supervised models, especially in lowlabel scenarios. The proposed method maintains the structural strengths of graph-based learning while resolving critical limitations associated with label homogenization. As a result, VPL offers a promising direction for future work in graph-structured SSL and neural graph learning.

## II. RELATED WORK

Graph-based semi-supervised learning (SSL) has been a pivotal area of research, offering frameworks that leverage both labeled and unlabeled data through structural graph information. Classical methods, such as Label Propagation (LP) [20], utilize the smoothness assumption to diffuse labels across the graph based on connectivity. Extensions like the Random Walk approach [21] and the Laplacian Regularization method [22] further embed manifold geometry into label inference.

Another notable advancement is the P-Laplacian-based framework, which generalizes the Laplace operator to better control diffusion by introducing nonlinearity through the $p$ parameter [23]. It enhances edge-aware smoothing but suffers from computational inefficiencies in large-scale settings. To address this, the Poisson learning model has emerged, applying a probabilistic viewpoint by aligning solutions with label constraints using divergence-based optimization [24].

Deep learning techniques have also found significant success in graph learning, especially with Graph Convolutional Networks (GCNs) [25] and Graph Attention Networks (GATs) [26]. These models integrate node features and topology for powerful end-to-end learning but often require more labeled data and suffer from over-smoothing in deeper architectures. Recent contributions, such as the Graph Poisson Network (GPN) [27], bridge classical PDE-based models with neural graph learning, enabling interpretable and stable predictions with fewer labeled nodes.

Variance-aware learning mechanisms have recently gained traction, motivated by the idea of adapting regularization strength to uncertainty in data regions. Work on adaptive Laplacian frameworks [28] and class-conditional regularization [29] have attempted to modulate diffusion intensity. However, these techniques often lack mathematical rigor or scalability. Our approach contributes to this line of research by developing variance-enhancing versions of Poisson and Laplace models, which dynamically amplify regularization based on label variance, improving robustness in low-label regimes.

Additionally, visual learning datasets such as FashionMNIST and CIFAR-10 [30], [31] are widely adopted to benchmark SSL methods, providing a diverse set of structured classification problems. Our evaluation also includes node classification on citation networks like Cora and CiteSeer [32], which are standard testbeds for graph neural network evaluation.

In summary, while previous methods have explored various graph diffusion and learning strategies, our proposed variancebased methods integrate label uncertainty directly into the learning process, offering a mathematically grounded and experimentally validated advancement in graph SSL.

## III. PRELIMINARIES

We consider a compact domain $\Omega \subseteq \mathbb{R}^d$, over which a probability distribution $\mu(x)$ is defined. The goal is to approximate a target function $u : \Omega \rightarrow Y$ that maps input data points to a label space Y. In the context of a multi-class classification problem, we define $Y = \{e_1,...,e_k\}$, where each $e_i$ is a one-hot encoded vector corresponding to a class label.

Assume that we are given a training dataset consisting of two subsets: a labeled set $\{(x_1,y_1),....,(x_l,y_l)\}$ and an unlabeled set $\{x_{l+1},...,x_n\}$, with all samples $x_i$ independently and identically drawn from the distribution $\mu(x)$.

To facilitate label propagation, we define an undirected weighted graph $G = (V,E)$ over the combined dataset $X = \{x_1,x_2,...,x_n\}$, where each node corresponds to a data point and the edge weights $w_{ij} \geq 0$ capture pairwise similarities (e.g., based on Euclidean distance or kernel-based affinities). The degree of a node $x_i$ is computed as $d_i = \sum_{j=1}^{n} w_{ij}$, representing the total connectivity strength of node $x_i$ to all other nodes.

In classical graph-based semi-supervised learning, the objective is to propagate label information from the small set of labeled nodes to the entire graph while respecting its structural connectivity. This is typically achieved by solving an equation involving the unnormalized graph Laplacian:

$$Lu(x_i) = 0, \qquad \text{for } i > l, \tag{2}$$

where the graph Laplacian operator is defined as

$$Lu(x_i) = \sum_{j=1}^{n} w_{ij}(u(x_i) - u(x_j)) \tag{3}$$

In the standard Laplacian learning approach, the labels for known data points are clamped by setting $u(x_i) = y_i$ for $i \leq l$. This enforces label consistency at the labeled nodes. An alternative and more flexible formulation is Poisson learning, where the graph Laplacian is augmented with a source term derived from the label deviation:

$$Lu(x_i) = y_i - \bar{y}, \text{ for } i \leq l, \tag{4} \text{ with the average}$$

label vector given by $\bar{y} = \frac{1}{l} \sum_{j=1}^{l} y_j$.

This modification allows the model to incorporate global information from the label distribution, enabling improved propagation behavior in sparse-label settings.

## IV. METHODOLOGY

### A. Variance-Enhanced Framework

Traditional graph-based semi-supervised learning techniques, such as those built on Laplacian or Poisson formulations, rely heavily on the assumption of label smoothness across the graph. Although effective in many cases, these techniques often fail in scenarios with limited labeled data, producing degenerate solutions where most nodes share nearly identical label predictions. This issue arises due to an overemphasis on minimizing smoothness loss, which suppresses informative diversity in the label space.

To overcome this limitation, we propose a *variance­regularized learning framework* that augments the classical objective with an explicit term that encourages label diversity. The goal is to find a balance between smoothness and expressiveness, thereby preventing trivial or constant predictions.

The proposed objective function is formulated as:

$$\sum_{i,j} w_{ij} \mathrm{L}(u(x_i), u(x_j)) - \lambda \cdot \mathrm{Var}[u], \tag{5}$$

where $w_{ij}$ denotes the affinity weight between nodes $x_i$ and $x_j$, $\mathrm{L}(\cdot, \cdot)$ represents a general pairwise loss function enforcing label consistency over edges, and $\mathrm{Var}[u]$ is a variance term that explicitly promotes label differentiation across the graph.

The variance component is mathematically expressed as:

$$\mathrm{Var}[u] = \sum_i q_i \|u(x_i) - \bar{u}\|^2, \tag{6}$$

with the degree-weighted average label $\bar{u}$ defined by:

$$\bar{u} = \sum_i q_i u(x_i), \quad q_i = \frac{d_i}{\sum_j d_j}, \tag{7}$$

where $d_i$ is the degree of node $x_i$, and $q_i$ serves as a normalized degree-based weighting. This design gives greater influence to high-degree nodes when computing the graphwide average and contributes more to the variance term.

By incorporating this variance-regularizing mechanism, the framework discourages convergence toward uniform solutions and instead drives the learning algorithm to discover more informative and distributed label assignments. The effect is especially pronounced under label-scarce conditions, where traditional methods typically fail.

If we consider the squared Euclidean distance as the pairwise loss:

$$\mathcal{L}(u(x_i), u(x_j)) = \frac{1}{2} \|u(x_i) - u(x_j)\|^2, \tag{8}$$

then the optimality conditions for the unlabeled nodes, derived from the Euler-Lagrange formulation, become:

$$Lu(x_i) = \lambda q_i(u(x_i) - \bar{u}), \qquad \text{for } i > l, \tag{9}$$

where $L$ is the graph Laplacian and $l$ denotes the number of labeled nodes.

### B. Variance-Enhanced Laplace Learning (V-Laplace)

Building upon the variance-enhanced learning framework, we propose a modified variant of Laplace learning, named *Variance-Enhanced Laplace Learning* (V-Laplace). This formulation incorporates the variance term into the learning dynamics to produce a label assignment that is both smooth and sufficiently diverse.

The governin

## V. EXPERIMENTAL EVALUATION

This section presents a comprehensive analysis of the performance of our proposed variance-enhanced models across a range of benchmark datasets, under sparse label settings. We report both quantitative results and qualitative insights into the behavior of the models. Our evaluation is divided into three key parts: image classification under limited supervision, mathematical formulation with interpretation, and node classification on citation networks.

### A. FashionMNIST Results

Table I shows classification accuracy for different methods on the FashionMNIST dataset, under varying numbers of labeled instances per class (from 1 to 5). Each entry indicates the mean and standard deviation across 100 independent runs. Among the methods compared, the proposed V-Laplace and VPoisson consistently demonstrate superior performance, with V-Poisson achieving the highest accuracy across all label regimes. The improvements are more pronounced in low-label settings, highlighting the model's effectiveness in handling sparse supervision.

TABLE I: Classification results (%) of multiple techniques on FashionMNIST with various label counts per class. Each value reflects the mean and standard deviation over 100 repetitions. Bold entries indicate improved variants, while the highest value per column is underlined.

| Labels/Class | 1 | 2 | 3 | 4 | 5 |
|---|---|---|---|---|---|
| Laplace/LP | 17.0 (6.6) | 31.7 (10.0) | 43.3 (8.4) | 52.8 (6.9) | 59.3 (5.7) |
| Nearest Neighbor | 43.9 (4.3) | 49.6 (3.3) | 52.7 (3.0) | 55.0 (2.4) | 56.9 (2.7) |
| Random Walk | 57.1 (4.8) | 63.1 (4.0) | 66.3 (2.8) | 68.5 (2.5) | 70.1 (2.2) |
| MBO | 15.7 (4.1) | 20.1 (4.6) | 25.7 (4.9) | 30.7 (4.9) | 34.8 (4.3) |
| WNLL | 43.0 (7.6) | 58.6 (5.1) | 64.0 (3.4) | 67.1 (3.4) | 69.6 (2.7) |
| Centered Kernel | 36.6 (4.2) | 47.2 (4.4) | 53.5 (3.9) | 58.4 (3.3) | 61.6 (3.4) |
| Sparse LP | 14.0 (5.5) | 14.0 (4.0) | 14.5 (4.0) | 18.0 (5.9) | 16.2 (4.2) |
| P-Laplace | 52.1 (4.8) | 58.4 (3.7) | 62.0 (3.0) | 64.3 (2.5) | 66.0 (2.5) |
| Poisson | 60.4 (4.7) | 66.3 (4.0) | 68.9 (2.7) | 70.7 (2.4) | 72.2 (2.2) |
| V-Laplace | 60.6 (5.0) | 66.3 (4.2) | 69.2 (2.8) | 71.0 (2.4) | 72.6 (2.3) |
| V-Poisson | 61.3 (4.9) | 67.1 (4.2) | 69.7 (2.8) | 71.3 (2.7) | 72.9 (2.3) |

### B. Modified Differential Equation Formulation

To better understand the mathematical behavior of our model, we consider the governing partial differential equation (PDE)

and its implications. Let $f : \mathbb{R}^d \to \mathbb{R}^k$ be a function with divergence:

$$\text{div}\,(\nabla f) = \left( \sum_{i=1}^{d} \frac{\partial^2 f_1}{\partial x_i^2}, \ldots, \sum_{i=1}^{d} \frac{\partial^2 f_k}{\partial x_i^2} \right)^\top$$

Define the operator $D_p(\nabla u)$ as:

$$D_p(\nabla u) = \text{diag}\left( \|\nabla u_1(x)\|_2^{p-2}, \ldots, \|\nabla u_k(x)\|_2^{p-2} \right)$$

Incorporating the distribution $\mu$ with full support and a normalized density $q(x)$, the PDE governing V-Poisson becomes:

$$\text{div}\,(\nabla u(x)) + 2\nabla u(x) \cdot \nabla \log \mu(x) + (p-2)\Delta_\infty u(x) + \frac{\lambda q(x)}{p \mu^2(x)} D_p^{-1}(\nabla u(x)) u(x) = 0 \tag{10}$$

subject to the constraint $\int u(x) q(x) dx = 0$. The term $\Delta_\infty u$ is the infinity Laplacian, defined component-wise as:

$$\Delta_\infty u = \left( \frac{(\nabla u_1)^\top \nabla^2 u_1 \nabla u_1}{\|\nabla u_1\|_2^2}, \ldots, \frac{(\nabla u_k)^\top \nabla^2 u_k \nabla u_k}{\|\nabla u_k\|_2^2} \right)$$

For the case where $d = k = 1$ and $p = 2$, the PDE simplifies to a familiar second-order ODE:

$$2\mu^2(x)u''(x) + 4\mu(x)\mu'(x)u'(x) + \lambda q(x)u(x) = 0$$

When $\mu(x)$ is uniform and $q(x) = \mu(x)$, this further reduces to:

$$u''(x) + \lambda u(x) = 0$$

with solution:

$$u(x) = C_1 \cos(\sqrt{\lambda} x) + C_2 \sin(\sqrt{\lambda} x)$$

### C. Visual Interpretation of Variance Enhancement

As illustrated in Fig. III, the classical Laplacian interpolation yields a harmonic, linear transition between labeled data points. However, when introducing variance-based scaling through the V-Poisson method, the solution adapts to data density. In areas of higher uncertainty or sparse labels, the regularization term intensifies, thereby reducing prediction variance and improving robustness—especially in multimodal distributions.

### D. CIFAR-10 Performance

To evaluate generalizability, we applied the methods to the CIFAR-10 dataset under similar sparse label settings. Table II summarizes the results. V-Poisson again surpasses baseline and prior methods, particularly excelling in extreme low-label regimes. This further validates the model's strength in realworld semi-supervised settings.

TABLE II: Classification accuracy (%) on CIFAR-10 with different label quantities per class.

| Method | 1 | 2 | 3 | 4 | 5 |
|---|---|---|---|---|---|
| Laplace/LP | 10.3 | 10.8 | 11.8 | 13.0 | 13.1 |
| Nearest Neighbor | 29.4 | 33.4 | 35.1 | 36.4 | 37.4 |
| Random Walk | 37.5 | 44.6 | 48.4 | 51.1 | 52.8 |
| MBO | 14.2 | 19.3 | 24.3 | 28.5 | 33.5 |
| WNLL | 14.9 | 24.9 | 33.2 | 38.4 | 42.4 |
| Centered Kernel | 35.6 | 42.7 | 46.0 | 48.6 | 50.1 |
| Sparse LP | 11.8 | 12.3 | 11.1 | 14.4 | 11.0 |
| P-Laplace | 34.7 | 41.3 | 44.6 | 47.2 | 48.8 |
| Poisson | 39.1 | 45.4 | 48.5 | 51.2 | 52.9 |
| V-Laplace | 33.9 | 40.5 | 44.0 | 46.6 | 47.8 |
| V-Poisson | 41.4 | 48.5 | 51.7 | 54.7 | 56.3 |

### E. Node Classification on Graph Data

Finally, we extend our approach to graph-based semisupervised learning. We benchmark the Variance-Enhanced Graph Poisson Network (V-GPN) against standard baselines (GCN, GAT, and GPN) on two citation networks: Cora and CiteSeer. Results in Table III clearly show that V-GPN consistently delivers higher accuracy, particularly in sparse label settings. These results substantiate the model's effectiveness across both grid and graph domains.

TABLE III: Node classification performance (%) with different label densities on Cora and CiteSeer datasets.

| Dataset | Method | 1 | 2 | 3 | 4 | 5 |
|---|---|---|---|---|---|---|
| Cora | GCN | 51.1 | 61.1 | 65.6 | 69.2 | 71.3 |
| | GAT | 50.5 | 59.7 | 63.1 | 66.7 | 68.8 |
| | GPN | 53.2 | 61.3 | 63.5 | 64.1 | 65.8 |
| | V-GPN | 58.9 | 64.1 | 67.6 | 70.5 | 72.8 |
| CiteSeer | GCN | 43.8 | 51.4 | 56.2 | 59.5 | 61.2 |
| | GAT | 42.0 | 51.1 | 53.6 | 58.7 | 58.5 |
| | GPN | 39.8 | 46.6 | 47.4 | 46.3 | 48.9 |
| | V-GPN | 49.5 | 55.4 | 57.6 | 62.1 | 63.2 |

## VI. RESULTS AND DISCUSSION

To validate the effectiveness of the proposed V-Laplace and V-Poisson formulations, we conducted extensive experiments on benchmark image datasets (FashionMNIST, CIFAR-10) and citation networks (Cora, CiteSeer). Performance was measured across a range of label-per-class settings, reflecting scenarios of extreme label scarcity to moderate availability. All accuracy scores reflect averages over 100 trials, with standard deviations reported to demonstrate consistency.

### A. FashionMNIST Results

Table I presents classification accuracy across different methods on the FashionMNIST dataset. The V-Poisson model consistently outperformed all baselines, especially under minimal label conditions. For instance, with just one labeled example per class, V-Poisson achieved an accuracy of 61.3%, surpassing both classical Poisson (60.4%) and advanced graphbased approaches such as Random Walk and P-Laplace.

Interestingly, the V-Laplace model also delivered competitive results, closely trailing the V-Poisson performance. The incorporation of label uncertainty via variance weighting effectively amplified the label propagation capability in lowlabel environments.

### B. CIFAR-10 Results

The more complex CIFAR-10 dataset, which presents greater visual variability and intra-class ambiguity, served as a further benchmark. As shown in Table II, our varianceenhanced models again demonstrated clear superiority under sparse supervision. Notably, V-Poisson reached an accuracy of 56.3% with five labels per class, a margin of more than 3% over the nearest contender, the classical Poisson method.

These results highlight the robustness of the varianceaware regularization term, which adapts well even in visually challenging settings, helping prevent over-smoothing and enhancing class separation.

### C. Graph Node Classification Results

We further evaluated our method in graph-based node classification tasks using the Cora and CiteSeer citation networks. Table III summarizes the accuracy scores for GCN, GAT, GPN, and the proposed V-GPN.

The V-GPN consistently outperformed all baselines across all label settings. For example, on Cora with only one label per class, V-GPN achieved 58.9% accuracy, outperforming GCN (51.1%) and GAT (50.5%). On the more sparse and challenging CiteSeer dataset, V-GPN reached 63.2% accuracy with five labels per class, showcasing its capability to generalize across graph structures and label distributions.

### D. Interpretation of Performance Gains

The consistent improvements observed in all tasks can be attributed to the following factors:

- The variance-enhancing mechanism increases sensitivity to uncertainty, allowing the model to better allocate confidence in label propagation.
- The adaptive regularization based on the underlying data density ($\mu(x)$) prevents oversmoothing and encourages sharper decision boundaries.
- The incorporation of higher-order differential operators like $\Delta_2 u$ contributes to more stable convergence and improved robustness in sparse-label scenarios.

### E. Visual Intuition

Figure IV-B provides qualitative insight into the behavior of V-Laplace under varying conditions. With increasing $\lambda$ and non-uniform data distributions, the model selectively enhances prediction certainty in ambiguous zones. The harmonic nature of classical Laplace solutions is preserved, while the variance-weighted formulation introduces beneficial gradientdriven nonlinearity that improves class separability.

## VII. CONCLUSION AND FUTURE WORK

In this work, we introduced a variance-amplified framework for semi-supervised learning on graphs and manifolds, aiming to tackle the inherent challenges posed by label sparsity. By integrating a variance-scaled regularization mechanism into classical Laplace and Poisson models, we derived the novel *VLaplace* and *V-Poisson* formulations. These enhanced models exhibited superior performance on benchmark image datasets (e.g., FashionMNIST, CIFAR-10) and graph-structured data (e.g., Cora, CiteSeer), especially under extreme low-label regimes.

The analytical derivation of the modified PDEs, grounded in variational principles and functional analysis, revealed how the incorporation of density-driven scaling and higher-order regularization fosters more confident and stable label propagation in uncertain regions. Theoretical observations were supported by extensive empirical validation across multiple domains. Furthermore, the proposed V-GPN architecture generalized this idea to deep neural graph networks, outperforming popular baselines such as GCN, GAT, and the original GPN in node classification tasks.

Future Work: Several promising directions remain open. First, the variance-aware framework can be extended to dynamic or temporal graphs where the topology evolves over time. Second, incorporating feature-level noise modeling could further refine confidence estimates during label propagation. Third, investigating scalable solvers and approximations for high-dimensional manifolds or large-scale graphs would make the proposed techniques applicable to real-time applications, such as social network analysis or autonomous navigation. Lastly, integrating this paradigm into end-to-end graph neural network training pipelines, possibly via joint optimization of feature embeddings and label diffusion parameters, could further elevate performance in more complex multimodal settings. Overall, our study underscores the critical role of uncertainty-aware mechanisms in enhancing graph-based learning and lays the foundation for future advances in labelefficient AI systems.